\pdfoutput=1

\documentclass[11pt]{article}

\input{def.set}
\usepackage{EMNLP2022}

\usepackage{times}
\usepackage{latexsym}

\usepackage[T1]{fontenc}

\usepackage[utf8]{inputenc}

\usepackage{microtype}

\usepackage{inconsolata}

\usepackage{amsmath}
\usepackage{amsfonts}
\usepackage{mathtools, cuted}
\usepackage{bm}
\usepackage{amssymb}
\usepackage{graphicx}
\usepackage{subcaption}
\usepackage{booktabs}
\usepackage{tikz}
\usepackage{pgfplots}
\usepackage{filecontents}
\usepackage{pifont}
\usepgfplotslibrary{fillbetween}
\usetikzlibrary{shapes,positioning,decorations.pathreplacing,shapes.multipart,arrows,chains,calc,plotmarks}
\usepackage{colortbl}

\newtheorem{theorem}{Theorem}[section]

\usepackage{enumitem}
\setenumerate[1]{itemsep=0pt,partopsep=0pt,parsep=\parskip,topsep=0pt}
\setitemize[1]{itemsep=2pt,partopsep=2pt,parsep=\parskip,topsep=2pt}
\setdescription{itemsep=0pt,partopsep=0pt,parsep=\parskip,topsep=0pt}

\newcommand*\circled[1]{\tikz[baseline=(char.base)]{
            \node[shape=circle,draw,inner sep=1pt,scale=0.8] (char) {#1};}}

\title{DORE: Document Ordered Relation Extraction based on Generative Framework}

\author{
Qipeng Guo\textsuperscript{2}\thanks{\ \ Equal contribution.} , 
Yuqing Yang\textsuperscript{1}\footnotemark[1] \thanks{\ \ Work done during internship at Amazon Shanghai AI Lab.} , 
Hang Yan\textsuperscript{1}, 
Xipeng Qiu\textsuperscript{1}\thanks{\ \ Corresponding author.} , 
Zheng Zhang\textsuperscript{2}  \\
\textsuperscript{1}School of Computer Science, Fudan University \\
\textsuperscript{2}Amazon AWS AI \\
\texttt{\{gqipeng, zhaz\}@amazon.com} \\
\texttt{yuqingyang21@m.fudan.edu.cn}, \texttt{\{hyan19, xpqiu\}@fudan.edu.cn} \\}

\begin{document}
\maketitle
\begin{abstract}
In recent years, there is a surge of generation-based information extraction work, which allows a more direct use of pre-trained language models and efficiently captures output dependencies. However, previous generative methods using lexical representation do not naturally fit document-level relation extraction (DocRE) where there are multiple entities and relational facts. In this paper, we investigate the root cause of the underwhelming performance of the existing generative DocRE models and discover that the culprit is the inadequacy of the training paradigm, instead of the capacities of the models. We propose to generate a symbolic and ordered sequence from the relation matrix which is deterministic and easier for model to learn. Moreover, we design a parallel row generation method to process overlong target sequences. 
Besides, we introduce several negative sampling strategies to improve the performance with balanced signals. Experimental results on four datasets show that our proposed method can improve the performance of the generative DocRE models. We have released our code at \url{https://github.com/ayyyq/DORE}.
\end{abstract}

\section{Introduction}
\begin{figure*}[t]
    \centering
    \includegraphics[width=1.0\linewidth]{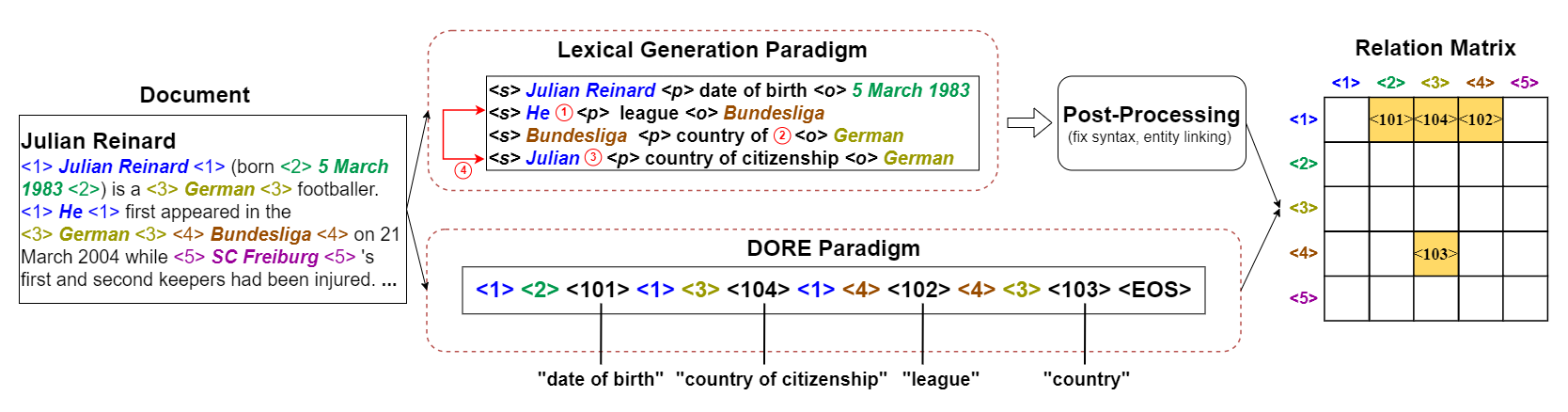}
    \caption{\small An example from DocRED dataset, and we highlight entities with different colors. The left side is the input document and the right side is the target relation matrix. Each cell in this matrix is filled with the relation between two entities. We compare two paradigms in the middle, and the lexical sequence is much longer than our DORE. Meanwhile, there are four weaknesses in lexical generation paradigm. \textcolor[rgb]{1.00,0.00,0.00}{\circled{1}} shows the case of using different mentions. Both ``\textcolor[rgb]{0.00,0.00,1.00}{Julian Reinard}'' and ``\textcolor[rgb]{0.00,0.00,1.00}{He}'' point to the same entity, but only one of them are used in the annotation. And this leads to an incorrect training signal if the model use a different mention. In \textcolor[rgb]{1.00,0.00,0.00}{\circled{2}}, the model struggles to choose from two similar relations ``\textcolor[RGB]{255,193,7}{country}'' and ``\textcolor[RGB]{255,193,7}{country of citizenship}''. The ``\textcolor[RGB]{255,193,7}{country of}'' is a meaningful lexicon but it is not valid in the relation vocabulary. In \textcolor[rgb]{1.00,0.00,0.00}{\circled{3}}, the model outputs a new mention ``\textcolor[rgb]{0.00,0.00,1.00}{Julian}'' that does not appear in the text. \textcolor[rgb]{1.00,0.00,0.00}{\circled{4}} shows that the prediction order does not follow the human reading order, as the knowledge of \textcolor[RGB]{0,150,136}{citizenship} appears before the \textcolor[RGB]{0,150,136}{league} information according to the document. Lexical generation paradigm adopts a post-processing step to address above issues. In contrast, our DORE paradigm directly predicts elements in the relation matrix.}
    \label{fig:top_ex}
\end{figure*}

Document-level relation extraction (DocRE) is a fundamental information extraction (IE) task which aims to extract relational facts among entities across multiple sentences. For IE, most previous approaches are classification-based, which first extract features of certain objects using pre-trained language models and then classify according to the merged features. Recent years have witnessed a rising trend of regarding the task of IE as a sequence generation problem, linearizing the extracted structures as a sequence. Compared to classification-based methods, generative framework extracts features and classifies simultaneously, allowing a more direct use of latent knowledge in pre-trained language models without an untrained classification module on the top. Besides, the generation process can naturally recover high-order dependencies when generating the output step by step. Generation-based methods have been successfully adapted to many settings including universal IE which intends to solve several IE tasks in a unified way \citep{DBLP:conf/iclr/PaoliniAKMAASXS21,DBLP:conf/acl/0001LDXLHSW22}, low-resource \citep{DBLP:journals/corr/abs-2108-12724}, and transfer learning \citep{DBLP:conf/emnlp/0030TCL021,DBLP:conf/acl/LiuHSW22}, and have achieved competitive results on most sentence-level benchmarks \citep{DBLP:conf/acl/CuiWLYZ21,DBLP:conf/emnlp/0030TCL021} and document-level event extraction task \citep{DBLP:conf/naacl/LiJH21,DBLP:conf/acl/Zhang0TW022}.

Considering that generative framework is simple and effective, prior work adopts it for DocRE \citep{DBLP:journals/corr/abs-2109-04901,DBLP:conf/bionlp/GiorgiBW22}. This line of work features \textit{lexical generation} since they use natural language to represent entities and relations, which directly borrows from text generation tasks \citep{DBLP:conf/acl/LewisLGGMLSZ20,DBLP:journals/jmlr/RaffelSRLNMZLL20}. Also, they need special separator tokens to distinguish token spans. However, the lexical generation paradigm dose not perfectly fit the more complex DocRE task, where the source sequence contains numerous entities and relations (e.g., a document can contain up to about 40 relation instances on DocRED), leading to a performance gap between generation-based and classification-based methods. Our experiment verifies that the generative baseline performs 6.00 
points worse than classification-based methods on DocRED dataset \citep{DBLP:conf/acl/YaoYLHLLLHZS19}.

The lexical generation paradigm faces two significant challenges, which impede its performance. (1) \textbf{non-unique target sequence}: for the example shown in Figure~\ref{fig:top_ex}, a document often needs to mention the same knowledge many times, each of which could be represented in multiple ways (i.e., diverse lexical forms for an entity). Pre-defining a certain lexical form of an entity will introduce incorrect bias, and a complicated post-processing step is needed to align the generated sequence and relational facts.
(2) \textbf{overlong generated sequence}: DocRE requires the model to extract more facts, leading to much longer output sequences, and thus causes difficulties to efficiency and memory support. 
However, it is hard for lexical generation approaches using natural language representation and extra separator tokens to cope with such dilemma in a concise way.

To alleviate issues in the lexical generation paradigm, we treat generative DocRE as deterministically generating a relation matrix where each cell corresponds to an entity pair with pre-defined relation or no relation. The paradigm, which we call DORE (\textbf{D}cument \textbf{O}dered \textbf{R}elation \textbf{E}xtraction), assigns each entity and relation a special id and linearizes the relation matrix in the row-column order, resulting a symbolic ordered sequence. It is much easier to learn and control generation. 
In addition, the paradigm is able to resolve overlong output sequences in a concise way when generating rows of the relation matrix in parallel. Besides, we show that the loss function taken from previous work 
is imbalanced for the complicated DocRE, and we introduce several negative sampling strategies to mitigate it. Taken together, we find that the underwhelming performance of generative framework for DocRE comes from the improper training and generation ways instead of the model architecture.

We conduct experiments on four popular DocRE benchmarks. We improve the generative model's $F_1$ score from 51.36 to 60.67 for DocRED by changing training paradigm only, and further improve it to 65.26 with distantly supervised training data (officially collected by DocRED). Besides, we bridge the performance gap between generation-based and classification-based methods on CDR \citep{DBLP:journals/biodb/LiSJSWLDMWL16} and GDA \citep{DBLP:conf/recomb/WuLLTL19} by obtaining 72.6 and 85.3 $F_1$ score, individually. We also achieve new state-of-the-art results on SciREX \citep{DBLP:conf/acl/JainZHB20} both with gold inputs and end-to-end for binary and 4-ary relation extraction. Our work brings generative framework to DocRE into a performance region that matches classification-based approaches, with the added advantage of supporting high-order relation discovery afforded by the nature of sequence-to-sequence models.


\section{Related Work} 
\subsection{Generation-based Information Extraction}
In recent years, more and more work seeks to use a new generative paradigm to solve information extraction tasks. \citet{DBLP:conf/iclr/PaoliniAKMAASXS21,DBLP:conf/acl/Zhang0DBL20} transform IE tasks into translation between label-augmented texts, \citet{DBLP:conf/acl/YanGDGZQ20,DBLP:conf/acl/0001LXHTL0LC20,DBLP:journals/corr/abs-2109-04901,DBLP:conf/acl/Zhang0TW022} design a linearization schema with constrained decoding strategies, and \citet{DBLP:conf/naacl/LiJH21,DBLP:journals/corr/abs-2108-12724,DBLP:conf/acl/LiuHSW22} adopt template-based conditional generation. Though simple the paradigm seems, generation-based methods report competitive results especially on sentence-level benchmarks. However, previous methods can not scale to the document-level relation extraction task which requires to extract multiple facts, or perform worse than most classification-based methods.

\subsection{Classification-based Document-Level Relation Extraction}
Most previous work treats DocRE as a classification task, which typically breaks down the task into two stages, extracting the feature of entities followed by classifying the relation of every entity pair according to their features. More specifically, a stream of classification-based work introduces the graph structure on top of pre-trained representations to address long-term dependencies and multi-hop reasoning \citep{DBLP:conf/acl/NanGSL20,DBLP:conf/emnlp/WangHCS20,DBLP:conf/emnlp/ZengXCL20,DBLP:conf/acl/ZengWC21,DBLP:conf/acl/XuCZ21}. However, for long document, compared with keeping a graph representation and merging new relations parsed from new paragraphs, using, for example, a seq2seq model, is a more scalable approach, and a direction worth exploring.

Recent work enhances classification-based methods in different aspects. \citet{DBLP:conf/ijcai/ZhangCXDTCHSC21} tackles the problem of lacking high-order dependencies by introducing convolution on relation matrices to encourage interaction among relations. On the other hand, \citet{DBLP:conf/aaai/XuWLZM21,DBLP:journals/corr/abs-2109-12093} enrich the features by introducing linguistic knowledge or statistic information of entities. Another popular idea \citep{DBLP:journals/corr/abs-2008-12283,DBLP:journals/corr/abs-2106-08657} is to detect the evidence sentences before relation extraction. This line of work provides a strong guideline for relation extraction and reduces irrelevant contexts. Some of these ideas are complementary to DORE; our core idea is to understand how generative framework can regain its advantages in dealing with document-level relation extraction and high-order relation discovery.

\section{Method}
\subsection{Task Formulation}
Document-level relation extraction task aims to extract relational facts given a document $\bD$ and a set of entities $\bE$. Each entity $e_i$ is represented as the set of its coreferent mentions $\{e^j_i\}$ in the document, some of which have different natural language forms. Each of the extracted instances can be expressed as a tuple $(e_1, \dots, e_k, r)$, where $k$ is the number of participating entities, and $r$ is from a pre-defined set of relations. We focus on binary and 4-ary relation extraction, that is, $k = 2 \text{ or } 4$. 

Since relation instances in the document can naturally formulate a matrix, we frame the generative DocRE as generating a relation matrix. Take binary relation extraction as an example. As shown in Figure~\ref{fig:top_ex}, each cell $(i, j)$ in the relation matrix corresponds to an entity pair with head entity $e_i$ and tail entity $e_j$, and can be filled with a relation. Then the goal of DocRE changes to estimate a conditional probability $P(\bR|\bD,\bE)$, where $\bR \in \mathbb{R}=[0,1]^{|\bE|\times |\bE| \times C}$ is a 3D-matrix, and $C$ is the number of relation categories. In practice, the goal is to find the most possible relation matrix.
\begin{gather}
    \bR^* = \argmax_{\bR \in \mathbb{R}} P(\bR|\bD, \bE).
\end{gather}
To further model DocRE as a sequence generation problem, we introduce a variable $\bS \in \mathbb{S}$ to represent a sequence. We will discuss the choice of how to represent this sequence space $\mathbb{S}$ shortly.
\begin{align}
P(\bR| \bD ,\bE) = \frac{\sum_{\bS \in \mathbb{S}} P(\bR, \bS, \bD, \bE)} {P(\bD, \bE)},& \\
= \sum_{\bS \in \mathbb{S}} P(\bR | \bS, \bD, \bE) P(\bS | \bD, \bE).& \label{eq:base}
\end{align}
Clearly, this computation is intractable, unless it is costly to enumerate the sequence space.

\subsection{Symbolic and Ordered Sequence Representation of Relation Matrix}
\label{sec:symrepr}

In our context, all we need to do is to represent the relation matrix and linearize it as a sequence.

To represent the relation matrix, we assign each entity and relation a special symbol, or, id, at first. In a real scenario, an entity can occur multiple times in the document by mentions, and expressions may be a little different in natural language, such as aliases, abbreviations or acronyms. A special id assures a unique and unambiguous entity. Besides, there is no need to use separators to distinguish entities and relations which contain more than one tokens. As shown in Figure~\ref{fig:top_ex}, we use different ranges of ``<i>'' (``<extra\_id\_i>'' in implementation) to represent entities ($i \in [1, 100]$) and relations ($i \in [101, 200]$). Entities are arranged according to their first appearance in the document. The embeddings of entity ids are initialized with those of corresponding sequential numbers. For example, we use the embedding of ``1'' to initialize the embedding of ``<1>''. Similarly, the embeddings of relation ids are initialized with the meaning pooling of the embeddings of the corresponding natural languages. The initialization benefits the pre-trained generative models to learn the meaning of the special tokens, as shown in Appendix-\ref{sec:analysis-init}. In this way, a relation tuple of \textit{(Julian Reinard, 5 March 1983, date of birth)} can be represented as ``<1> <2> <101>'' in our paradigm.

For linearization of the relation matrix, we simply organize relation tuples in the row-column order. The result is that the relation instances whose head entity appears earlier in the document proceed those appear later in the output sequence, and the order of relations sharing the same head entity is decided by their tail entities. The optimal order that complies with the logical reasoning is hard to define in advance, unless the model bears heavy computation to enumerate the sequence space. On the contrary, the row-column order is deterministic and easy for the model to understand.

More formally, let $\hat{\bS}$ be the corresponding sequence of the relation matrix $\bR$, i.e., $\sum_{\bS \in \mathbb{S}} P(\bR, \bS) = P(\bR, \hat{\bS})$, and $P(\bR|\hat{\bS}, \bD, \bE) = P(\bR|\hat{\bS}) = 1$. Let  $\tau(\cdot)$ be the linearization function that converts a relation matrix to a sequence following the symbolic format and row-column order we described above, i.e., $ \hat{\bS}=\tau(\bR)$. We have:
\begin{align}
&P(\bR|\bD,\bE) \\
=& \sum_{\bS \in \mathbb{S}} P(\bR | \bS, \bD, \bE) P(\bS | \bD, \bE),\\
=& P(\bR|\hat{\bS},\bD,\bE) P(\hat{\bS} | \bD, \bE), \\
=& P(\hat{\bS} | \bD, \bE). \label{eq:seq2seq} 
\end{align}

\paragraph{4-ary Relation Extraction} The symbolic relation matrix can be easily extended to 4-ary relation extraction and the setting where entity type information is provided. For instance, a 4-ary relation instance $(e_{\text{Task}}, e_{\text{Method}}, e_{\text{Material}}, e_{\text{Metric}}, r)$ in SciREX is composed of four types of entities and a binary relation. Each type of entities can be further divided into different ranges of ``<i>'', and a relation tuple can be transformed to a similar sequence like ``<1> <26> <51> <76> <101>''. 

\paragraph{Constrained Decoding} Considering that the reference sequence is completely a series of triples for binary relation extraction or 5-ary tuples for 4-ary relation extraction, we utilize a relatively simple constrained decoding method to control generation compared to lexical generation paradigm. The vocabulary is confined to a certain range of special tokens barely depending on the current step, so that the decoder's vocabulary size is small. On the contrary, lexical generation methods requires a full vocabulary because they need to predict entities' text forms.

\subsection{Parallel Row Generation}
Due to the autoregressive nature of generative models, longer the output, slower the decoding process. Besides, when the output is too long, the memory is not supported. Fortunately, our method can easily accommodate to such situation. Instead of generating the whole relation matrix in one pass, we can choose to generate one row of the relation matrix each time. For example, in Figure~\ref{fig:top_ex}, the model first only generates relation triples started with ``\textit{Julian Reinard}'' which is denotes as ``<1>'' in the output sequence, and then restarts to generate other rows of the relation matrix in turn in the same way. Since the input is the same, the procedure can be parallel, thus saving time and generated length. All we need is different decoder start tokens. The parallel row generation sacrifices some relation dependencies, but saves length. And experiments in Sec-\ref{sec:results-docred} show that the trade-off is positive.

\subsection{Loss Function Design}
\label{sec:loss}

The relation matrix is typically sparse for DocRE task. For example, there are only approximately 3\% entity pairs having relations on DocRED. Some previous work has found that sampling negative training examples, that is, entity pairs having no relation, during training is effective to improve the model performance. We propose several negative sampling strategies for our method and explain the reason in terms of loss function.
 
Our training target is the ground truth sequence $\bS^*=(i,j,\bR_{ij})_{i,j \in \bR^+}$. Here, $\bR^+$ is a set of all nonzero elements in the relation matrix, $i$ is the row index and $j$ is the column index according to the row-column order. Similarly, we denote $\bR^-$ as the set of all zero elements. This produces the following generative loss function:
\begin{gather}
\begin{split}
    \mathcal{L}_{\text{seq}} =& \sum_{t=1}^{T-1} \text{CE}({\bS^*}_t, P(\bS_t|\bS_{<t},\bD,\bE)) \\ +& \text{CE}(\text{<EOS>}, \bS_T),
\end{split}
 \label{eq:loss_f}
\end{gather}
where $T$ is the sequence length, and the last token is ``<EOS>'', which means the end of sequence. 

However, the loss function to generate the relation matrix is applying cross entropy \citep{cox1958regression} to each element in the relation matrix:
\begin{align}
    \mathcal{L} &= \sum_{i,j} \text{CE}(\bR_{ij}, P(\bR_{ij}|\bD,\bE)), \\
\begin{split}
&=\sum_{i,j \in \bR^+} \text{CE}(\bR_{ij}, P(\bR_{ij}|\bD,\bE)) \\
&+\sum_{i,j \in \bR^-} \text{CE}(\bR_{ij}, P(\bR_{ij}|\bD,\bE)).  \label{eq:loss_two_sets}
\end{split} 
\end{align}
By comparison, the sequence generation loss which we use in practice purges all the zero elements from the $\bR^-$ set and lumps all of their effect into predicting the end of sequence. While it certainly shortens the target sequence, it also leads to severe imbalance loss terms. On the other hand, having the generative model produces all of zero elements defeats the purpose of symbolic format and can easily exceed the maximum length supported by most pre-trained seq2seq models. What is needed here is to balance the population of negative samples without overwhelming the generative model.

\begin{figure}[t]
\centering
\begin{subfigure}{.48\columnwidth}
  \centering
  \includegraphics[width=.6\linewidth]{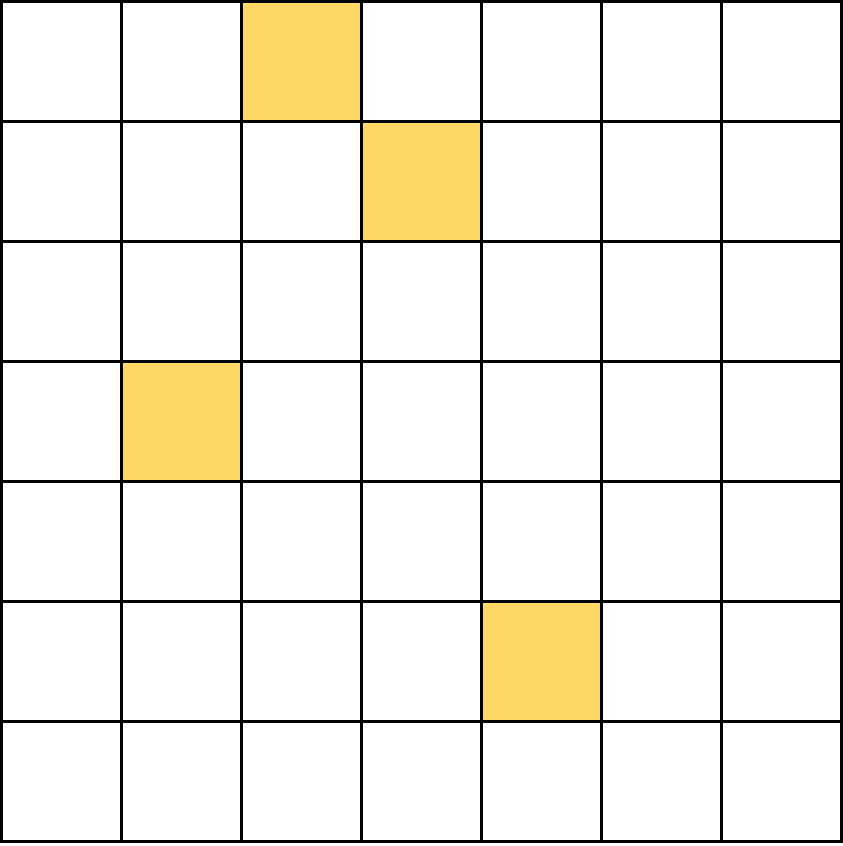}
  \caption{no sampling}
  \label{fig:sfig1}
\end{subfigure}
\begin{subfigure}{.48\columnwidth}
  \centering
  \includegraphics[width=.6\linewidth]{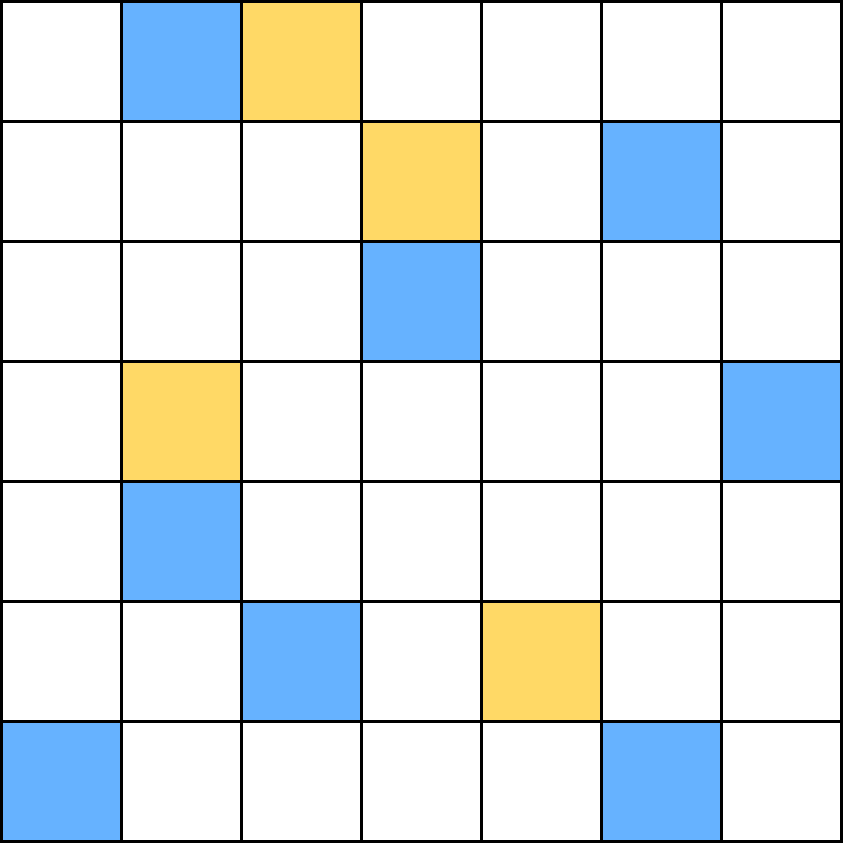}
  \caption{random sampling}
  \label{fig:sfig2}
\end{subfigure} \\
\begin{subfigure}{.48\columnwidth}
  \centering
  \includegraphics[width=.6\linewidth]{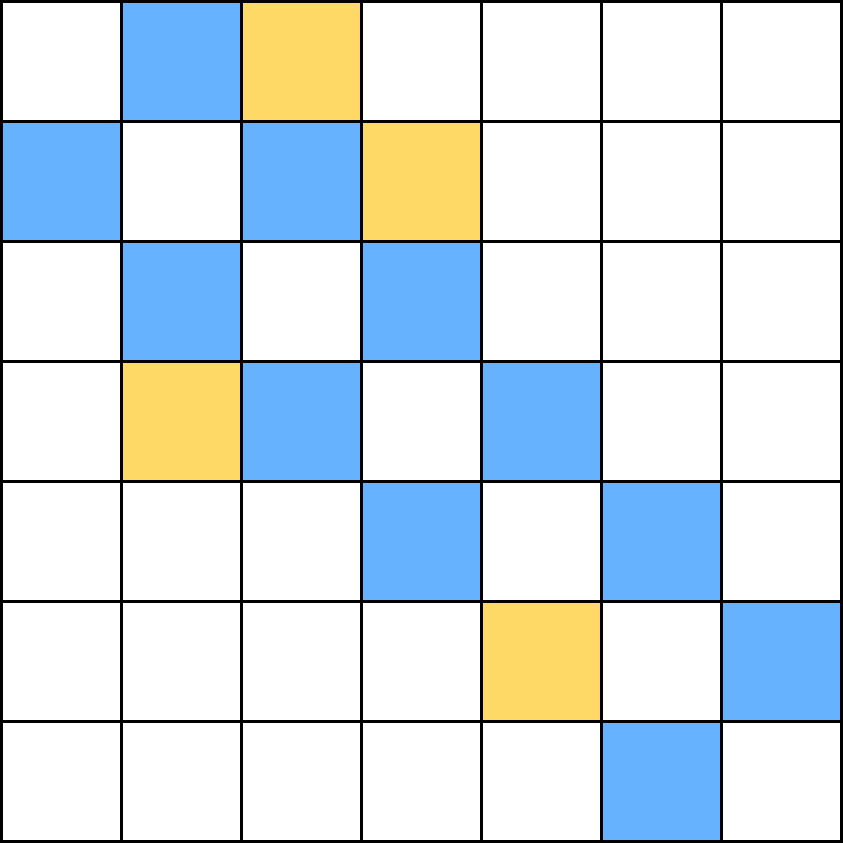}
  \caption{diagonal sampling}
  \label{fig:sfig3}
\end{subfigure}
\begin{subfigure}{.48\columnwidth}
  \centering
  \includegraphics[width=.6\linewidth]{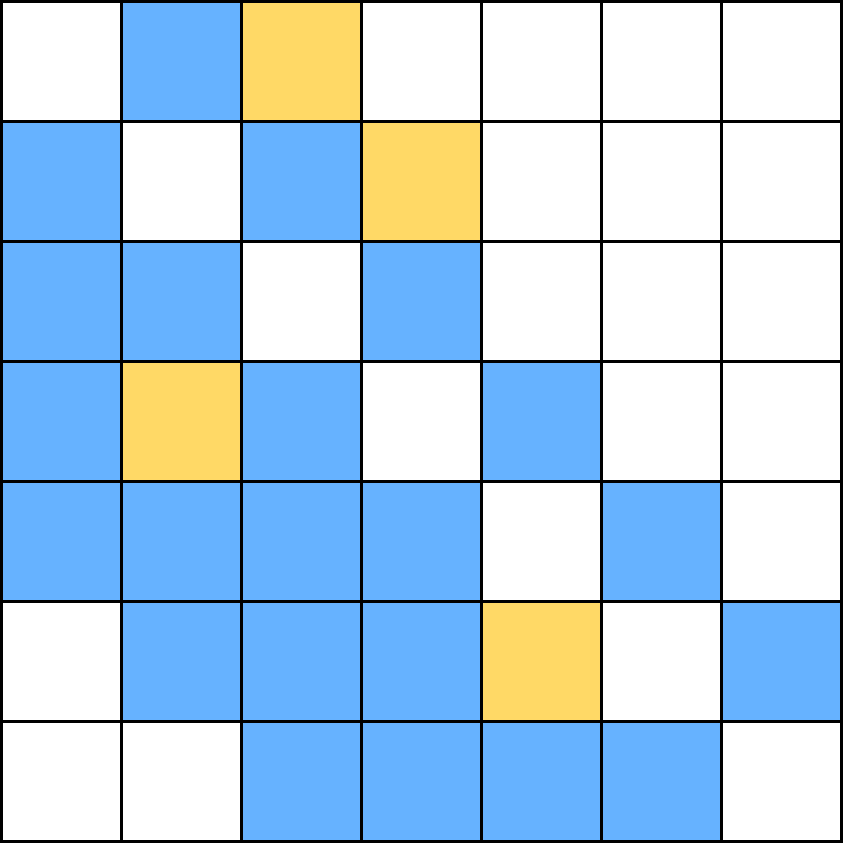}
  \caption{asymmetric sampling}
  \label{fig:sfig4}
\end{subfigure}
\caption{\small Different strategies of negative sampling. Light yellow elements are annotated relations, and blue elements are negative samples. There is a flipped version of (d) used in the dynamic sampling, which keeps one element in the left side of the diagonal and multiple elements in the right side. }
\label{fig:fig}
\end{figure}

\paragraph{Negative Sampling Strategies} Straightforwardly, we can add random zero elements in the relation matrix as negative samples, as shown in Figure~\ref{fig:sfig2}. For example, we randomly pick 10\% zero elements and add them to the target sequence. However, this raises the difficulty of sequence prediction since the model might struggle to remember the order for each training sample, ignoring the contextual information. A more effective way is to regularly preserve elements in the diagonal band with a constant (and therefore balanced) budget of the relation matrix, namely, \textit{diagonal negative sampling}. Further to alleviate the bias brought by the limited nonzero element space, we introduce a dynamic strategy to provide negative samples by randomly picking from: no sampling (Figure~\ref{fig:sfig1}), diagonal sampling (Figure~\ref{fig:sfig3}), and asymmetric sampling (Figure~\ref{fig:sfig4}). We call it \textit{dynamic negative sampling}. In evaluation, we remove the zero elements the model generates.

\section{Experiments}

\subsection{Datasets and Evaluation Metrics}
\begin{table}[!htbp]
\centering
\small
\begin{tabular}{lccc}
    \toprule
    \bf Dataset & \bf \# Train & \bf \# Dev & \bf \# Test \\
    \midrule
    DocRED & 3053 & 1000 & 1000 \\
    CDR & 500 & 500 & 500 \\
    GDA & 23353 & 5839 & 1000 \\
    SciREX & 306 & 66 & 66 \\
    \bottomrule
\end{tabular}
\caption{\small Statistics of the datasets in experiments.}
\label{tab:dataset}
\end{table}

\begin{table*}[!htbp]
\centering
\small
\begin{tabular}{lcccc}
    \toprule
     & \multicolumn{2}{c}{\bf Dev} & \multicolumn{2}{c}{\bf Test} \\
    \bf Model & \bf Ign $\bf F_1$ & $\bf F_1$ & \bf Ign $\bf F_1$ & $\bf F_1$ \\
    \midrule
    \multicolumn{5}{l}{\textbf{\textit{classification-based}}} \\
    BERT$_{\textrm{base}}$ ~\citep{DBLP:journals/corr/abs-1909-11898} & - & 54.16 & - & 53.20 \\
    \rowcolor{gray!30} T5$_{\textrm{large}}$ & 56.20 & 57.99 & 55.44 & 57.36 \\
    RoBERTa$_{\textrm{large}}$ \citep{DBLP:conf/emnlp/YeLDLLSL20} & 57.19 & 59.40 & 57.74 & 60.06 \\
    strong RoBERTa$_{\textrm{large}}$ \citep{DBLP:conf/aaai/XuWLZM21} & 58.45 & 60.58 & 58.43 & 60.54 \\
    SAIS$^{\textrm{B}}_{\textrm{All}}$-RoBERTa$_{\textrm{large}}$ \citep{DBLP:journals/corr/abs-2109-12093} & $62.23 \pm 0.15$ & $65.17 \pm 0.08$ & 63.44 & 65.11 \\
    NCRL+ATLOP-DeBERTa$_\textrm{large}$ + distant $\dagger$ \citep{DBLP:journals/corr/abs-2205-00476} & $\underline{66.11 \pm 0.14}$ & $\underline{67.92 \pm 0.14}$ & $\underline{65.81}$ & $\underline{67.53}$ \\
    \midrule
    \multicolumn{5}{l}{\textbf{\textit{generation-based}}} \\
    \rowcolor{gray!30} lexical generation & 48.43 & 50.34 & 49.32 & 51.36 \\
    DORE & 52.79 & 55.12 & 52.53 & 55.10 \\
    DORE + negative sampling$_{\text{dynamic}}$ & 58.43 & 60.42 & 57.58 & 59.88 \\
    DORE + negative sampling$_{\text{all}}$ + parallel row generation & $58.55 \pm 0.11$ & $60.61 \pm 0.10$ & 58.44 & 60.67 \\
    DORE + negative sampling$_{\text{diagonal}}$ + distant $\dagger$  & $\bf 62.91 \pm 0.13$ & $\bf 64.70 \pm 0.12$ & \bf 63.26 & \bf 65.26 \\
    \bottomrule
\end{tabular}
\caption{\small Main results on DocRED. Results with $\dagger$ mean the models are pre-trained on the distantly supervised dataset provided by DocRED. Rows in gray denote the models are implemented by ourselves. The best results are underlined and the best results of the generation-based models are in bold.}
\label{tab:main-results-docred}
\end{table*}

We evaluate our model on four commonly used DocRE datasets. \textbf{DocRED} \citep{DBLP:conf/acl/YaoYLHLLLHZS19} is a human-annotated DocRE dataset with 96 relation types between two entities. Articles and their relation sets are mined from Wikipedia. \textbf{CDR} \citep{DBLP:journals/biodb/LiSJSWLDMWL16} is a manually annotated dataset for DocRE in the biochemical domain. The aim is to predict whether there is a chemical-induced disease (CID) relation between Chemical and Disease. \textbf{GDA} \citep{DBLP:conf/recomb/WuLLTL19} is also a biochemical dataset annotated with binary interactions between Gene and Disease concepts at the document-level via distant supervision. \textbf{SciREX} \citep{DBLP:conf/acl/JainZHB20} is a document-level information extraction dataset, including binary and 4-ary relation extraction from scientific articles. It contains four types of entities, Task, Method, Material, and Metric, and coreference is annotated. The dataset statistics are listed in Table~\ref{tab:dataset}.

For DocRED, we use $F_1$ and Ign $F_1$ in evaluation following \citet{DBLP:conf/acl/YaoYLHLLLHZS19}, where Ign $F_1$ denotes $F_1$ removing triples having appeared in both the training and development/testing set. For other datasets, we use $F_1$ in evaluation.

\subsection{Implementation details}
We implement our model in PyTorch, and use pre-trained generative models provided by huggingface \footnote{https://huggingface.co/} as the backbone. For DocRED, we choose T5 \citep{DBLP:journals/jmlr/RaffelSRLNMZLL20}, which is pre-trained on a multi-task mixture of unsupervised and supervised tasks and shows power in a variety of NLP tasks. In fact, any pre-trained generative models can be used, and we show the experiments in Appendix-\ref{sec:backbone}. For the two biochemical datasets, we use BioBART \citep{DBLP:conf/bionlp/YuanYGZXY22}, which adapts BART \citep{DBLP:conf/acl/LewisLGGMLSZ20} to the biochemical domain and benefits the domain-specific sequence generation tasks. For SciREX, we choose Longformer-Encoder-Decoder (LED) \citep{DBLP:journals/corr/abs-2004-05150} as the backbone, a Longformer variant for supporting long document seq2seq tasks, since documents in SciREX are much longer than 1024. The models are trained using the AdamW \citep{DBLP:conf/iclr/LoshchilovH19} optimizer with weight decay coefficient of 0.01, and a linearly decaying scheduler \citep{DBLP:journals/corr/GoyalDGNWKTJH17}. Other hyperparameters are listed in Appendix-\ref{sec:appendix-hyperparam}. All experiments are conducted with Tesla T4 GPUs.

\subsection{Main Results}

\subsubsection{Comparison on DocRED}
\label{sec:results-docred}

In Table~\ref{tab:main-results-docred}, ``T5$_{\textrm{large}}$'' is a classification baseline that we replace the ``Enhanced BERT Baseline'' using a bilinear classifier on top of the pre-trained language model provided by \citet{DBLP:conf/aaai/Zhou0M021} with T5$_{\text{large}}$'s encoder. All generation-based methods are implemented by ourselves using T5$_{\text{large}}$. ``lexical generation'' means that we directly adopt T5 under lexical generation paradigm. ``DORE'' refers to the symbolic and ordered sequence representation for relation matrix introduced in Sec-\ref{sec:symrepr}. ``+ negative sampling'' and ``+ parallel row generation'' add negative sampling and parallel row generation, respectively. Finally, ``distant'' means that the model is first trained on the additional noisy distant training corpus (provided by DocRED) and then fine-tuned on the human-annotated training set.

Experiments using T5 demonstrate that our proposed framework is effective. Results of ``DORE'' show that adopting the symbolic and ordered sequence format improves the test $F_1$ score against the lexical generation by 3.74 points. Adding dynamic negative sampling further brings 4.78 points improvement, which intuitively verifies the imbalance training signal harms the classification task and mitigating it brings significant benefit. When using all nonzero elements in the relation matrix and utilizing parallel row generation to support it with limited resources, there is still a minor improvement. It proves that the proposed parallel row generation method is practical, and considering more negative samples is valuable with less data. 
Pre-training on the distant corpus, we adopt diagonal negative sampling to reduce the computational cost, and can achieve 65.26 test $F_1$ on DocRED \footnote{Experiment results show that adopting diagonal negative sampling is enough when the model is pre-trained on the large-scale distantly supervised dataset.}.

Besides generation-based methods, we also compare DORE with classification-based methods. The most effective setting ``DORE + negative sampling$_{\text{all}}$ + parallel row generation'', abbreviated as ``DORE$_{\textrm{NS + PRG}}$'', improves ``T5$_{\textrm{large}}$'' by 3.31 points, demonstrating that DORE can beat the classification-based method with the same feature extraction flow (using T5) since the only difference between these two methods is how they compute the relation matrix. Also, ``DORE$_{\textrm{NS + PRG}}$'' outperforms BERT$_{\textrm{base}}$ and RoBERTa$_\textrm{large}$ baselines without changing the model architecture or using extra training data. Admittedly, our proposed method still has a performance gap with SOTA methods on DocRED, which employ a series of advanced techniques. For example, \textbf{SAIS$^{\textrm{B}}_{\textrm{All}}$-RoBERTa$_{\textrm{large}}$} designed complicated pipeline multi-task learning and data augmentation, and \textbf{NCRL+ATLOP-DeBERTa$_\textrm{large}$ + distant} utilized DeBERTa$_\textrm{large}$ which is proved to be more powerful than RoBERTa$_\textrm{large}$ on DocRED \citep{DBLP:journals/corr/abs-2205-00476}. In contrast, our generative framework is more concise and potential. 

\subsubsection{Comparison on CDR and GDA}
\begin{table}[!htbp]
\centering
\small
\resizebox{\columnwidth}{!}{
\begin{tabular}{lcc}
    \toprule
    \bf Model & \bf CDR & \bf GDA \\
    \midrule
    \multicolumn{3}{l}{\textbf{\textit{classification-based}}} \\
    EoG \citep{DBLP:conf/emnlp/ChristopoulouMA19} & 63.6 & 81.5 \\
    \rowcolor{gray!30} BioBART$_{\text{base}}$ & 64.1 & 81.6 \\
    SciBERT \citep{DBLP:conf/aaai/Zhou0M021} & 65.1 & 82.5 \\
    \rowcolor{gray!30} BioBART$_{\text{large}}$ & 67.3 & 82.3 \\
    SSAN-SciBERT \citep{DBLP:conf/aaai/XuWLZM21} & 68.7 & 83.7 \\
    ATLOP-SciBERT \citep{DBLP:conf/aaai/Zhou0M021} & 69.4 & 83.9 \\
    DocuNet-SciBERT \citep{DBLP:conf/ijcai/ZhangCXDTCHSC21} & 76.3 & 85.3 \\
    SAIS$^{\textrm{O}}_{\textrm{RE+CR+ET}}$-SciBERT \citep{DBLP:journals/corr/abs-2109-12093} & $\underline{79.0}$ & $\underline{87.1}$ \\
    \midrule
    \multicolumn{3}{l}{\textbf{\textit{generation-based}}} \\
    seq2rel \citep{DBLP:conf/bionlp/GiorgiBW22} & 67.2 & 84.9 \\
    DORE-BioBART$_{\text{base}}$ & 69.0 & 84.7 \\
    DORE-BioBART$_{\text{large}}$ & \textbf{72.6} & \textbf{85.3} \\
    \bottomrule
\end{tabular}}
\caption{\small Test $F_1$ scores on CDR and GDA. Row in gray denote the models are implemented by ourselves.}
\label{tab:main-results-bio}
\end{table}

For a fair comparison, we leverage entity type information when evaluating on the two biochemical datasets. ``BioBART$_{\text{base}}$'' and ``BioBART$_{\text{large}}$'' are classification baselines by replacing ``SciBERT'' implemented by \citet{DBLP:conf/aaai/Zhou0M021} with corresponding BioBART's encoder. \textbf{Seq2rel} is a lexical generation method and employs copy mechanism and entity hinting to control generation.

As shown in Table~\ref{tab:main-results-bio}, our method improves the performance of the previous generative DocRED method on CDR using BioBART$_{\text{base}}$ and BioBART$_{\text{large}}$, and further bridges the gap between classification-based and generation-based methods. It illustrates the advantage of symbolic sequence representation. DORE-BioBART$_{\text{base}}$ performs slightly worse than seq2rel on GDA, and we owe it to the weakness of BioBART on this dataset, given the comparison between classification-based methods using SciBERT and BioBART. Besides, DORE outperforms corresponding classification-based methods by 4.9/3.1 and 5.3/3.0 points on CDR/GDA  using BioBART$_{\text{base}}$ and BioBART$_{\text{large}}$, respectively, which verifies the strength of our generative framework.

\subsubsection{Comparison on SciREX}
\begin{table}[!htbp]
\centering
\small
\resizebox{\columnwidth}{!}{
\begin{tabular}{lcccccc}
    \toprule
     & \multicolumn{3}{c}{\bf Binary RE} & \multicolumn{3}{c}{\bf 4-ary RE} \\
    \bf Model & \bf P & \bf R & $\bf F_1$ & \bf P & \bf R & $\bf F_1$ \\
    \midrule
    \multicolumn{7}{c}{\bf Component-wise (gold input)} \\
    \midrule
    SciREX-P & 82.0 & 44.0 & 57.0 & 53.1 & 71.8 & 61.1 \\
    DORE-LED$_{\textrm{base}}$ & 88.7 & 77.8 & \textbf{82.9} & 79.5 & 55.5 & \textbf{65.4} \\
    \midrule
    \multicolumn{7}{c}{\bf End-to-end} \\
    \midrule
    TANL-BART$_{\textrm{base}}$ & 0.74 & 0.67 & 0.62 & 0.00 & 0.00 & 0.00 \\
    D{\small Y}GIE++ & 2.9 & 12.8 & 3.8 & - & - & - \\
    SciREX-P & 6.5 & 44.1 & 9.6 & 0.7 & 17.3 & 0.8 \\
    TempGen-BART$_{\textrm{base}}$ & 17.11 & 13.56 & 14.47 & 3.19 & 4.26 & 3.55 \\
    TempGen-LED$_{\textrm{base}}$$^*$ & 18.75 & 14.48 & 15.59 & 0.00 & 0.00 & 0.00 \\
    DORE-LED$_{\textrm{base}}$ & 30.45 & 23.93 & \textbf{26.80} & 9.52 & 5.41 & \textbf{6.90} \\
    \bottomrule
\end{tabular}}
\caption{\small Main results on SciREX. Results with $*$ denote the models are implemented by ourselves.}
\label{tab:main-results-scirex}
\end{table}

In Table~\ref{tab:main-results-scirex}, \textbf{D{\small Y}GIE++} \citep{DBLP:conf/emnlp/WaddenWLH19} and \textbf{SciREX-P} \citep{DBLP:conf/acl/JainZHB20} are classification-based methods, while \textbf{TANL} \citep{DBLP:conf/iclr/PaoliniAKMAASXS21} and \textbf{TempGen-BART$_{\text{base}}$} \citep{DBLP:journals/corr/abs-2109-04901} are lexical generation-based methods in general. We replace BART$_{\text{base}}$ with LED$_{\text{base}}$ for TempGen, namely, TempGen-LED$_{\text{base}}$. There is a slight improvement using TempGen-LED$_{\text{base}}$ on the binary relation extraction, mainly because encoding longer documents (4096 vs. 1024) provides more useful contextual information and relational facts. However, it cannot extract valid entities for 4-ary relation extraction. We attribute the bad performance to lexical generation paradigm making the model confused to represent entities.

To fairly compare when evaluating using gold inputs, we add entity type information. To compare with end-to-end relation extraction methods, we adopt \textit{fast-coref} \footnote{\url{https://github.com/shtoshni/fast-coref}} \citep{DBLP:conf/emnlp/ToshniwalWELG20} to resolve coreference resolution using Longformer$_{\text{base}}$, which achieves 34.5 $F_1$ score while D{\small Y}GIE++ 47.6 and SciREX-P 25.5. Experiments show that our proposed method achieves new SOTA results on both binary and 4-ary relation extraction tasks in both settings, which demonstrates the effectiveness and generalization of DORE.

\subsection{Ablation Study}
\label{sec:ablation}

\paragraph{Symbolic vs. Lexical}
In this section, we compare the lexical representation and our symbolic representation. We choose T5$_{\textrm{large}}$ as the testbed, which can learn the symbolic representation without constrained decoding.

The upper part of Table~\ref{tab:ex_ablation} shows that the symbolic representation betters the performance by 3.76 points, which is a substantial improvement. We believe the improvement comes from two aspects: the symbolic representation largely reduces the sequence length, alleviating the accumulation decoding error; and it simplifies the copy mechanism since one symbol represents a long text phrase.

\paragraph{Sequence Order}
The sequence order plays an essential role in DORE. To understand its effect, we compare the annotation order, i.e., the order of how annotators annotate a document, and the row-column order, against a reference baseline using random order, where each sample is associated with a random sequence to be generated. Experiments are conducted with T5$_{\textrm{large}}$ and symbolic formatted sequences.

\begin{table}[t]
    \centering\small
    \begin{tabular}{ccc}
    \toprule
      \bf Method & \bf Ign $\bf F_1$ & $\bf F_1$\\
      \midrule
         lexical & 48.43 & 50.34 \\
         symbolic & \bf 52.02 & \bf 54.10 \\
        \midrule
         random order & 51.40 & 53.39 \\
         annotation order & 52.02 & 54.10 \\
         row-column order & \bf 52.79 & \bf 55.12 \\
        \midrule
         10\% random & 52.89 & 55.45 \\
         diagonal & 56.75 & 58.81 \\
         dynamic & \bf 58.55 & \bf 60.61 \\
        \bottomrule
    \end{tabular}
    \caption{\small Ablation studies of symbolic representation, sequence order, and negative sampling. All results are from DocRED dev set. The best results in each block are in bold.}
    \label{tab:ex_ablation}
\end{table}

The middle part of Table~\ref{tab:ex_ablation} gives a comparison between three orders. The annotation order does outperform the random order since it is more predictable. However, we can not assume that annotators' behaviors are consistent. As we expected, the row-column order, which is not only stable but also deterministic, further outperforms the annotation order by 1.02 points. Still, we do not believe it is necessarily the best order. In general, a better order should reflect high-order dependencies' topology, and we leave this as a future direction.

\paragraph{Negative Sampling}

We also test different negative sampling strategies introduced in Sec-\ref{sec:loss}. There are three settings. ``10\% random'' uniformly picks 10\% negative samples from the relation matrix. ``diagonal'' means we select negative samples with window size of 1 around the diagonal. And ``dynamic'' uniformly selects the different strategies we introduced before. We test these settings with T5$_{\textrm{large}}$ plus symbolic representation and row-column order. We found the ``10\% random'' option contributes little, the ``diagonal'' outperforms it since it is consistent across entities. Finally, the ``dynamic'' option performs the best because it provides the model a chance to see all negative samples in different passes.

\subsection{High-order Dependencies}

\begin{figure}
    \centering
    \includegraphics[width=0.8\linewidth]{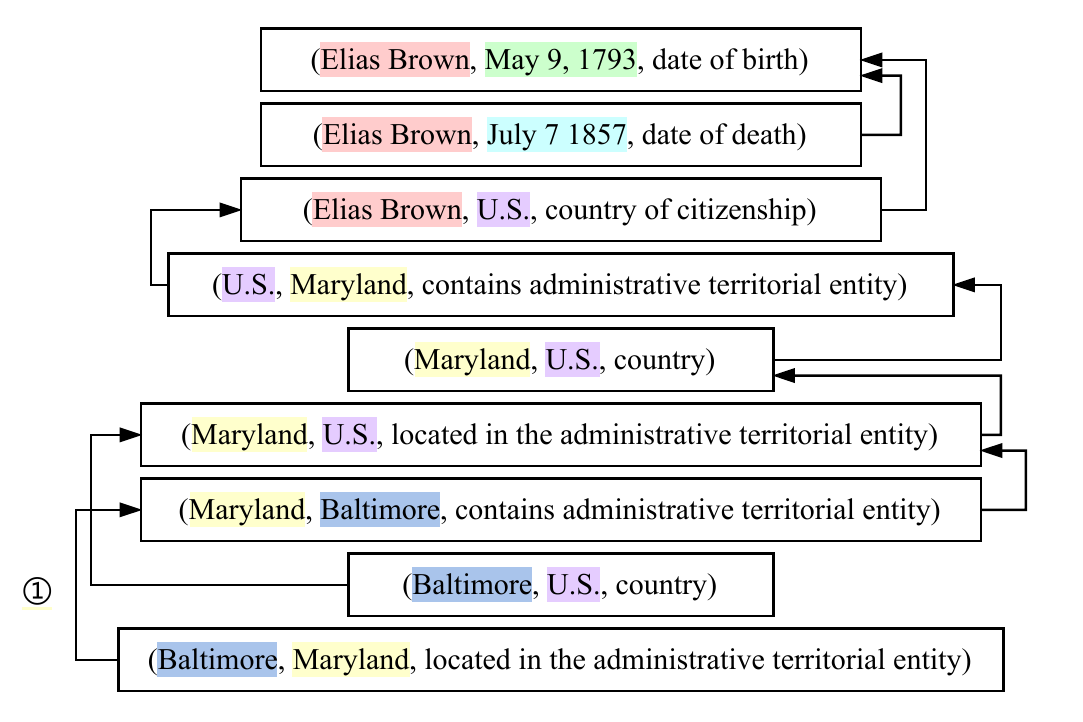}
    \caption{\small Case study of how DORE attends to previously generated triples from the document ``Elias Brown (May 9,1793 – July 7, 1857) was a U.S. Representative from Maryland. Born near Baltimore, Maryland, Brown attended the common schools …'' }
    \label{fig:vis}
\end{figure}
To verify whether the proposed model captures high-order dependencies, we provide a case study in Figure~\ref{fig:vis} by probing into decoder attention scores. For each triple, we draw an edge to the generated triples in previous steps that receive highest attention. And we explain how to compute this score in Appendix-\ref{sec:appendix-vis}. 
 
The decoder would always attend to the last generated triples if it could not recover output dependencies. In contrast, we can find that the decoder of DORE tends to attend to the previous triples with the same head or tail entity according to Figure~\ref{fig:vis}, which is more likely to bed latent associations. As the example \circled{1} shows, the latter triple accurately predicts the symmetric relation based on the former one. Conventional classification-based methods can not do this without additional modules. 

\section{Conclusion}
We propose a new generative paradigm DORE for DocRE. DORE adopts a symbolic and ordered sequence representation, establishing a clean connection between the sequence generation and DocRE. We also introduce parallel row generation and several negative sampling methods to improve the effectiveness and efficiency. Experiments on four DocRE datasets demonstrate that our method can substantially improve generative models without changing their designs.


\section*{Limitations}
As shown in Table~\ref{tab:main-results-docred} and Table~\ref{tab:main-results-bio}, although our proposed method without extra modules has outperformed classification baselines which have a simple classifier on top of pre-trained language models, there still exists a performance gap on the relatively complicated DocRED and domain-specific CDR and GDA. For one thing, we assume that some techniques proposed by SOTA work are complementary for DORE, and experiments are needed to verify whether DORE faces same issues. For another, experiments show that encoders of pre-trained generative models including T5 and BioBART are weaker to extract features compared to popular non-generative models like RoBERTa and DeBERTa, which impedes the model performance to some extent. Therefore, replacing the generative model's encoder with pre-trained language models used for classification for a fairer comparison leaves for future work.

\section*{Ethics Statement}
Our work complies with the ACL Ethics Policy. As document-level relation extraction is a standard task in NLP, and all datasets we used are public, we do not see any critical ethical considerations.

\section*{Acknowledgement}
We would like to express gratitude to the anonymous reviewers for their kind and insightful comments. This work was supported by the National Key Research and Development Program of China (No.2020AAA0108700) and National Natural Science Foundation of China (No.62022027).

\bibliography{custom}

\begin{thebibliography}{41}
\expandafter\ifx\csname natexlab\endcsname\relax\def\natexlab#1{#1}\fi

\bibitem[{Beltagy et~al.(2020)Beltagy, Peters, and
  Cohan}]{DBLP:journals/corr/abs-2004-05150}
Iz~Beltagy, Matthew~E. Peters, and Arman Cohan. 2020.
\newblock \href {http://arxiv.org/abs/2004.05150} {Longformer: The
  long-document transformer}.
\newblock \emph{CoRR}, abs/2004.05150.

\bibitem[{Christopoulou et~al.(2019)Christopoulou, Miwa, and
  Ananiadou}]{DBLP:conf/emnlp/ChristopoulouMA19}
Fenia Christopoulou, Makoto Miwa, and Sophia Ananiadou. 2019.
\newblock \href {https://doi.org/10.18653/v1/D19-1498} {Connecting the dots:
  Document-level neural relation extraction with edge-oriented graphs}.
\newblock In \emph{Proceedings of the 2019 Conference on Empirical Methods in
  Natural Language Processing and the 9th International Joint Conference on
  Natural Language Processing, {EMNLP-IJCNLP} 2019, Hong Kong, China, November
  3-7, 2019}, pages 4924--4935. Association for Computational Linguistics.

\bibitem[{Cox(1958)}]{cox1958regression}
David~R Cox. 1958.
\newblock The regression analysis of binary sequences.
\newblock \emph{Journal of the Royal Statistical Society: Series B
  (Methodological)}, 20(2):215--232.

\bibitem[{Cui et~al.(2021)Cui, Wu, Liu, Yang, and
  Zhang}]{DBLP:conf/acl/CuiWLYZ21}
Leyang Cui, Yu~Wu, Jian Liu, Sen Yang, and Yue Zhang. 2021.
\newblock \href {https://doi.org/10.18653/v1/2021.findings-acl.161}
  {Template-based named entity recognition using {BART}}.
\newblock In \emph{Findings of the Association for Computational Linguistics:
  {ACL/IJCNLP} 2021, Online Event, August 1-6, 2021}, volume {ACL/IJCNLP} 2021
  of \emph{Findings of {ACL}}, pages 1835--1845. Association for Computational
  Linguistics.

\bibitem[{Giorgi et~al.(2022)Giorgi, Bader, and
  Wang}]{DBLP:conf/bionlp/GiorgiBW22}
John~M. Giorgi, Gary~D. Bader, and Bo~Wang. 2022.
\newblock \href {https://aclanthology.org/2022.bionlp-1.2} {A
  sequence-to-sequence approach for document-level relation extraction}.
\newblock In \emph{Proceedings of the 21st Workshop on Biomedical Language
  Processing, BioNLP@ACL 2022, Dublin, Ireland, May 26, 2022}, pages 10--25.
  Association for Computational Linguistics.

\bibitem[{Goyal et~al.(2017)Goyal, Doll{\'{a}}r, Girshick, Noordhuis,
  Wesolowski, Kyrola, Tulloch, Jia, and
  He}]{DBLP:journals/corr/GoyalDGNWKTJH17}
Priya Goyal, Piotr Doll{\'{a}}r, Ross~B. Girshick, Pieter Noordhuis, Lukasz
  Wesolowski, Aapo Kyrola, Andrew Tulloch, Yangqing Jia, and Kaiming He. 2017.
\newblock \href {http://arxiv.org/abs/1706.02677} {Accurate, large minibatch
  {SGD:} training imagenet in 1 hour}.
\newblock \emph{CoRR}, abs/1706.02677.

\bibitem[{Hsu et~al.(2021)Hsu, Huang, Boschee, Miller, Natarajan, Chang, and
  Peng}]{DBLP:journals/corr/abs-2108-12724}
I{-}Hung Hsu, Kuan{-}Hao Huang, Elizabeth Boschee, Scott Miller, Prem
  Natarajan, Kai{-}Wei Chang, and Nanyun Peng. 2021.
\newblock \href {http://arxiv.org/abs/2108.12724} {Event extraction as natural
  language generation}.
\newblock \emph{CoRR}, abs/2108.12724.

\bibitem[{Huang et~al.(2020)Huang, Wang, Ma, and
  Huang}]{DBLP:journals/corr/abs-2008-12283}
Kevin Huang, Guangtao Wang, Tengyu Ma, and Jing Huang. 2020.
\newblock Entity and evidence guided relation extraction for docred.
\newblock \emph{CoRR}, abs/2008.12283.

\bibitem[{Huang et~al.(2021)Huang, Tang, and
  Peng}]{DBLP:journals/corr/abs-2109-04901}
Kung{-}Hsiang Huang, Sam Tang, and Nanyun Peng. 2021.
\newblock Document-level entity-based extraction as template generation.
\newblock \emph{CoRR}, abs/2109.04901.

\bibitem[{Jain et~al.(2020)Jain, van Zuylen, Hajishirzi, and
  Beltagy}]{DBLP:conf/acl/JainZHB20}
Sarthak Jain, Madeleine van Zuylen, Hannaneh Hajishirzi, and Iz~Beltagy. 2020.
\newblock \href {https://doi.org/10.18653/v1/2020.acl-main.670} {Scirex: {A}
  challenge dataset for document-level information extraction}.
\newblock In \emph{Proceedings of the 58th Annual Meeting of the Association
  for Computational Linguistics, {ACL} 2020, Online, July 5-10, 2020}, pages
  7506--7516. Association for Computational Linguistics.

\bibitem[{Lewis et~al.(2020)Lewis, Liu, Goyal, Ghazvininejad, Mohamed, Levy,
  Stoyanov, and Zettlemoyer}]{DBLP:conf/acl/LewisLGGMLSZ20}
Mike Lewis, Yinhan Liu, Naman Goyal, Marjan Ghazvininejad, Abdelrahman Mohamed,
  Omer Levy, Veselin Stoyanov, and Luke Zettlemoyer. 2020.
\newblock {BART:} denoising sequence-to-sequence pre-training for natural
  language generation, translation, and comprehension.
\newblock In \emph{{ACL}}, pages 7871--7880. Association for Computational
  Linguistics.

\bibitem[{Li et~al.(2016)Li, Sun, Johnson, Sciaky, Wei, Leaman, Davis,
  Mattingly, Wiegers, and Lu}]{DBLP:journals/biodb/LiSJSWLDMWL16}
Jiao Li, Yueping Sun, Robin~J. Johnson, Daniela Sciaky, Chih{-}Hsuan Wei,
  Robert Leaman, Allan~Peter Davis, Carolyn~J. Mattingly, Thomas~C. Wiegers,
  and Zhiyong Lu. 2016.
\newblock \href {https://doi.org/10.1093/database/baw068} {Biocreative {V}
  {CDR} task corpus: a resource for chemical disease relation extraction}.
\newblock \emph{Database J. Biol. Databases Curation}, 2016.

\bibitem[{Li et~al.(2021)Li, Ji, and Han}]{DBLP:conf/naacl/LiJH21}
Sha Li, Heng Ji, and Jiawei Han. 2021.
\newblock \href {https://doi.org/10.18653/v1/2021.naacl-main.69}
  {Document-level event argument extraction by conditional generation}.
\newblock In \emph{Proceedings of the 2021 Conference of the North American
  Chapter of the Association for Computational Linguistics: Human Language
  Technologies, {NAACL-HLT} 2021, Online, June 6-11, 2021}, pages 894--908.
  Association for Computational Linguistics.

\bibitem[{Liu et~al.(2021)Liu, Teng, Cui, Liu, and
  Zhang}]{DBLP:conf/emnlp/0030TCL021}
Jian Liu, Zhiyang Teng, Leyang Cui, Hanmeng Liu, and Yue Zhang. 2021.
\newblock \href {https://doi.org/10.18653/v1/2021.emnlp-main.361} {Solving
  aspect category sentiment analysis as a text generation task}.
\newblock In \emph{Proceedings of the 2021 Conference on Empirical Methods in
  Natural Language Processing, {EMNLP} 2021, Virtual Event / Punta Cana,
  Dominican Republic, 7-11 November, 2021}, pages 4406--4416. Association for
  Computational Linguistics.

\bibitem[{Liu et~al.(2022)Liu, Huang, Shi, and Wang}]{DBLP:conf/acl/LiuHSW22}
Xiao Liu, Heyan Huang, Ge~Shi, and Bo~Wang. 2022.
\newblock \href {https://aclanthology.org/2022.acl-long.358} {Dynamic
  prefix-tuning for generative template-based event extraction}.
\newblock In \emph{Proceedings of the 60th Annual Meeting of the Association
  for Computational Linguistics (Volume 1: Long Papers), {ACL} 2022, Dublin,
  Ireland, May 22-27, 2022}, pages 5216--5228. Association for Computational
  Linguistics.

\bibitem[{Loshchilov and Hutter(2019)}]{DBLP:conf/iclr/LoshchilovH19}
Ilya Loshchilov and Frank Hutter. 2019.
\newblock Decoupled weight decay regularization.
\newblock In \emph{{ICLR} (Poster)}. OpenReview.net.

\bibitem[{Lu et~al.(2021)Lu, Lin, Xu, Han, Tang, Li, Sun, Liao, and
  Chen}]{DBLP:conf/acl/0001LXHTL0LC20}
Yaojie Lu, Hongyu Lin, Jin Xu, Xianpei Han, Jialong Tang, Annan Li, Le~Sun,
  Meng Liao, and Shaoyi Chen. 2021.
\newblock \href {https://doi.org/10.18653/v1/2021.acl-long.217} {Text2event:
  Controllable sequence-to-structure generation for end-to-end event
  extraction}.
\newblock In \emph{Proceedings of the 59th Annual Meeting of the Association
  for Computational Linguistics and the 11th International Joint Conference on
  Natural Language Processing, {ACL/IJCNLP} 2021, (Volume 1: Long Papers),
  Virtual Event, August 1-6, 2021}, pages 2795--2806. Association for
  Computational Linguistics.

\bibitem[{Lu et~al.(2022)Lu, Liu, Dai, Xiao, Lin, Han, Sun, and
  Wu}]{DBLP:conf/acl/0001LDXLHSW22}
Yaojie Lu, Qing Liu, Dai Dai, Xinyan Xiao, Hongyu Lin, Xianpei Han, Le~Sun, and
  Hua Wu. 2022.
\newblock \href {https://aclanthology.org/2022.acl-long.395} {Unified structure
  generation for universal information extraction}.
\newblock In \emph{Proceedings of the 60th Annual Meeting of the Association
  for Computational Linguistics (Volume 1: Long Papers), {ACL} 2022, Dublin,
  Ireland, May 22-27, 2022}, pages 5755--5772. Association for Computational
  Linguistics.

\bibitem[{Nan et~al.(2020)Nan, Guo, Sekulic, and Lu}]{DBLP:conf/acl/NanGSL20}
Guoshun Nan, Zhijiang Guo, Ivan Sekulic, and Wei Lu. 2020.
\newblock Reasoning with latent structure refinement for document-level
  relation extraction.
\newblock In \emph{{ACL}}, pages 1546--1557. Association for Computational
  Linguistics.

\bibitem[{Paolini et~al.(2021)Paolini, Athiwaratkun, Krone, Ma, Achille,
  Anubhai, dos Santos, Xiang, and Soatto}]{DBLP:conf/iclr/PaoliniAKMAASXS21}
Giovanni Paolini, Ben Athiwaratkun, Jason Krone, Jie Ma, Alessandro Achille,
  Rishita Anubhai, C{\'{\i}}cero~Nogueira dos Santos, Bing Xiang, and Stefano
  Soatto. 2021.
\newblock Structured prediction as translation between augmented natural
  languages.
\newblock In \emph{{ICLR}}. OpenReview.net.

\bibitem[{Raffel et~al.(2020)Raffel, Shazeer, Roberts, Lee, Narang, Matena,
  Zhou, Li, and Liu}]{DBLP:journals/jmlr/RaffelSRLNMZLL20}
Colin Raffel, Noam Shazeer, Adam Roberts, Katherine Lee, Sharan Narang, Michael
  Matena, Yanqi Zhou, Wei Li, and Peter~J. Liu. 2020.
\newblock Exploring the limits of transfer learning with a unified text-to-text
  transformer.
\newblock \emph{J. Mach. Learn. Res.}, 21:140:1--140:67.

\bibitem[{Toshniwal et~al.(2020)Toshniwal, Wiseman, Ettinger, Livescu, and
  Gimpel}]{DBLP:conf/emnlp/ToshniwalWELG20}
Shubham Toshniwal, Sam Wiseman, Allyson Ettinger, Karen Livescu, and Kevin
  Gimpel. 2020.
\newblock \href {https://doi.org/10.18653/v1/2020.emnlp-main.685} {Learning to
  ignore: Long document coreference with bounded memory neural networks}.
\newblock In \emph{Proceedings of the 2020 Conference on Empirical Methods in
  Natural Language Processing, {EMNLP} 2020, Online, November 16-20, 2020},
  pages 8519--8526. Association for Computational Linguistics.

\bibitem[{Wadden et~al.(2019)Wadden, Wennberg, Luan, and
  Hajishirzi}]{DBLP:conf/emnlp/WaddenWLH19}
David Wadden, Ulme Wennberg, Yi~Luan, and Hannaneh Hajishirzi. 2019.
\newblock \href {https://doi.org/10.18653/v1/D19-1585} {Entity, relation, and
  event extraction with contextualized span representations}.
\newblock In \emph{Proceedings of the 2019 Conference on Empirical Methods in
  Natural Language Processing and the 9th International Joint Conference on
  Natural Language Processing, {EMNLP-IJCNLP} 2019, Hong Kong, China, November
  3-7, 2019}, pages 5783--5788. Association for Computational Linguistics.

\bibitem[{Wang et~al.(2020)Wang, Hu, Cao, and Sun}]{DBLP:conf/emnlp/WangHCS20}
Difeng Wang, Wei Hu, Ermei Cao, and Weijian Sun. 2020.
\newblock Global-to-local neural networks for document-level relation
  extraction.
\newblock In \emph{{EMNLP} {(1)}}, pages 3711--3721. Association for
  Computational Linguistics.

\bibitem[{Wang et~al.(2019)Wang, Focke, Sylvester, Mishra, and
  Wang}]{DBLP:journals/corr/abs-1909-11898}
Hong Wang, Christfried Focke, Rob Sylvester, Nilesh Mishra, and William~Yang
  Wang. 2019.
\newblock Fine-tune bert for docred with two-step process.
\newblock \emph{CoRR}, abs/1909.11898.

\bibitem[{Wu et~al.(2019)Wu, Luo, Leung, Ting, and
  Lam}]{DBLP:conf/recomb/WuLLTL19}
Ye~Wu, Ruibang Luo, Henry C.~M. Leung, Hing{-}Fung Ting, and Tak~Wah Lam. 2019.
\newblock \href {https://doi.org/10.1007/978-3-030-17083-7\_17} {{RENET:} {A}
  deep learning approach for extracting gene-disease associations from
  literature}.
\newblock In \emph{Research in Computational Molecular Biology - 23rd Annual
  International Conference, {RECOMB} 2019, Washington, DC, USA, May 5-8, 2019,
  Proceedings}, volume 11467 of \emph{Lecture Notes in Computer Science}, pages
  272--284. Springer.

\bibitem[{Xiao et~al.(2021)Xiao, Zhang, Mao, Yang, and
  Han}]{DBLP:journals/corr/abs-2109-12093}
Yuxin Xiao, Zecheng Zhang, Yuning Mao, Carl Yang, and Jiawei Han. 2021.
\newblock {SAIS:} supervising and augmenting intermediate steps for
  document-level relation extraction.
\newblock \emph{CoRR}, abs/2109.12093.

\bibitem[{Xie et~al.(2021)Xie, Shen, Li, Mao, and
  Han}]{DBLP:journals/corr/abs-2106-08657}
Yiqing Xie, Jiaming Shen, Sha Li, Yuning Mao, and Jiawei Han. 2021.
\newblock Eider: Evidence-enhanced document-level relation extraction.
\newblock \emph{CoRR}, abs/2106.08657.

\bibitem[{Xu et~al.(2021{\natexlab{a}})Xu, Wang, Lyu, Zhu, and
  Mao}]{DBLP:conf/aaai/XuWLZM21}
Benfeng Xu, Quan Wang, Yajuan Lyu, Yong Zhu, and Zhendong Mao.
  2021{\natexlab{a}}.
\newblock Entity structure within and throughout: Modeling mention dependencies
  for document-level relation extraction.
\newblock In \emph{{AAAI}}, pages 14149--14157. {AAAI} Press.

\bibitem[{Xu et~al.(2021{\natexlab{b}})Xu, Chen, and
  Zhao}]{DBLP:conf/acl/XuCZ21}
Wang Xu, Kehai Chen, and Tiejun Zhao. 2021{\natexlab{b}}.
\newblock Discriminative reasoning for document-level relation extraction.
\newblock In \emph{{ACL/IJCNLP} (Findings)}, volume {ACL/IJCNLP} 2021 of
  \emph{Findings of {ACL}}, pages 1653--1663. Association for Computational
  Linguistics.

\bibitem[{Yan et~al.(2021)Yan, Gui, Dai, Guo, Zhang, and
  Qiu}]{DBLP:conf/acl/YanGDGZQ20}
Hang Yan, Tao Gui, Junqi Dai, Qipeng Guo, Zheng Zhang, and Xipeng Qiu. 2021.
\newblock \href {https://doi.org/10.18653/v1/2021.acl-long.451} {A unified
  generative framework for various {NER} subtasks}.
\newblock In \emph{Proceedings of the 59th Annual Meeting of the Association
  for Computational Linguistics and the 11th International Joint Conference on
  Natural Language Processing, {ACL/IJCNLP} 2021, (Volume 1: Long Papers),
  Virtual Event, August 1-6, 2021}, pages 5808--5822. Association for
  Computational Linguistics.

\bibitem[{Yao et~al.(2019)Yao, Ye, Li, Han, Lin, Liu, Liu, Huang, Zhou, and
  Sun}]{DBLP:conf/acl/YaoYLHLLLHZS19}
Yuan Yao, Deming Ye, Peng Li, Xu~Han, Yankai Lin, Zhenghao Liu, Zhiyuan Liu,
  Lixin Huang, Jie Zhou, and Maosong Sun. 2019.
\newblock Docred: {A} large-scale document-level relation extraction dataset.
\newblock In \emph{{ACL} {(1)}}, pages 764--777. Association for Computational
  Linguistics.

\bibitem[{Ye et~al.(2020)Ye, Lin, Du, Liu, Li, Sun, and
  Liu}]{DBLP:conf/emnlp/YeLDLLSL20}
Deming Ye, Yankai Lin, Jiaju Du, Zhenghao Liu, Peng Li, Maosong Sun, and
  Zhiyuan Liu. 2020.
\newblock Coreferential reasoning learning for language representation.
\newblock In \emph{{EMNLP} {(1)}}, pages 7170--7186. Association for
  Computational Linguistics.

\bibitem[{Yuan et~al.(2022)Yuan, Yuan, Gan, Zhang, Xie, and
  Yu}]{DBLP:conf/bionlp/YuanYGZXY22}
Hongyi Yuan, Zheng Yuan, Ruyi Gan, Jiaxing Zhang, Yutao Xie, and Sheng Yu.
  2022.
\newblock \href {https://aclanthology.org/2022.bionlp-1.9} {Biobart:
  Pretraining and evaluation of {A} biomedical generative language model}.
\newblock In \emph{Proceedings of the 21st Workshop on Biomedical Language
  Processing, BioNLP@ACL 2022, Dublin, Ireland, May 26, 2022}, pages 97--109.
  Association for Computational Linguistics.

\bibitem[{Zeng et~al.(2021)Zeng, Wu, and Chang}]{DBLP:conf/acl/ZengWC21}
Shuang Zeng, Yuting Wu, and Baobao Chang. 2021.
\newblock {SIRE:} separate intra- and inter-sentential reasoning for
  document-level relation extraction.
\newblock In \emph{{ACL/IJCNLP} (Findings)}, volume {ACL/IJCNLP} 2021 of
  \emph{Findings of {ACL}}, pages 524--534. Association for Computational
  Linguistics.

\bibitem[{Zeng et~al.(2020)Zeng, Xu, Chang, and Li}]{DBLP:conf/emnlp/ZengXCL20}
Shuang Zeng, Runxin Xu, Baobao Chang, and Lei Li. 2020.
\newblock Double graph based reasoning for document-level relation extraction.
\newblock In \emph{{EMNLP} {(1)}}, pages 1630--1640. Association for
  Computational Linguistics.

\bibitem[{Zhang et~al.(2021{\natexlab{a}})Zhang, Chen, Xie, Deng, Tan, Chen,
  Huang, Si, and Chen}]{DBLP:conf/ijcai/ZhangCXDTCHSC21}
Ningyu Zhang, Xiang Chen, Xin Xie, Shumin Deng, Chuanqi Tan, Mosha Chen, Fei
  Huang, Luo Si, and Huajun Chen. 2021{\natexlab{a}}.
\newblock Document-level relation extraction as semantic segmentation.
\newblock In \emph{{IJCAI}}, pages 3999--4006. ijcai.org.

\bibitem[{Zhang et~al.(2022)Zhang, Shen, Tan, Wu, and
  Lu}]{DBLP:conf/acl/Zhang0TW022}
Shuai Zhang, Yongliang Shen, Zeqi Tan, Yiquan Wu, and Weiming Lu. 2022.
\newblock \href {https://aclanthology.org/2022.acl-long.59} {De-bias for
  generative extraction in unified {NER} task}.
\newblock In \emph{Proceedings of the 60th Annual Meeting of the Association
  for Computational Linguistics (Volume 1: Long Papers), {ACL} 2022, Dublin,
  Ireland, May 22-27, 2022}, pages 808--818. Association for Computational
  Linguistics.

\bibitem[{Zhang et~al.(2021{\natexlab{b}})Zhang, Li, Deng, Bing, and
  Lam}]{DBLP:conf/acl/Zhang0DBL20}
Wenxuan Zhang, Xin Li, Yang Deng, Lidong Bing, and Wai Lam. 2021{\natexlab{b}}.
\newblock \href {https://doi.org/10.18653/v1/2021.acl-short.64} {Towards
  generative aspect-based sentiment analysis}.
\newblock In \emph{Proceedings of the 59th Annual Meeting of the Association
  for Computational Linguistics and the 11th International Joint Conference on
  Natural Language Processing, {ACL/IJCNLP} 2021, (Volume 2: Short Papers),
  Virtual Event, August 1-6, 2021}, pages 504--510. Association for
  Computational Linguistics.

\bibitem[{Zhou et~al.(2021)Zhou, Huang, Ma, and
  Huang}]{DBLP:conf/aaai/Zhou0M021}
Wenxuan Zhou, Kevin Huang, Tengyu Ma, and Jing Huang. 2021.
\newblock Document-level relation extraction with adaptive thresholding and
  localized context pooling.
\newblock In \emph{{AAAI}}, pages 14612--14620. {AAAI} Press.

\bibitem[{Zhou and Lee(2022)}]{DBLP:journals/corr/abs-2205-00476}
Yang Zhou and Wee~Sun Lee. 2022.
\newblock \href {https://doi.org/10.48550/arXiv.2205.00476} {None class ranking
  loss for document-level relation extraction}.
\newblock \emph{CoRR}, abs/2205.00476.

\end{thebibliography}
\bibliographystyle{acl_natbib}

\appendix

\section{Appendix}
\label{sec:appendix}

\subsection{Initialization of Entity ID}
\label{sec:analysis-init}

\definecolor{deepblue}{RGB}{45,149,191}
\definecolor{deepred}{RGB}{241,90,90}

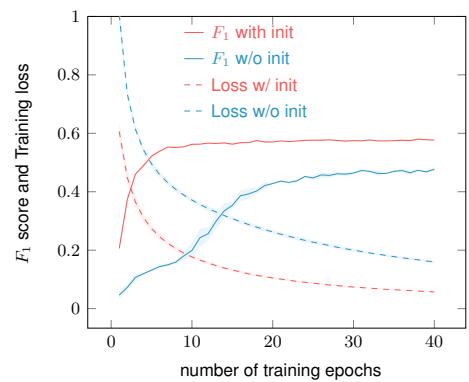
\begin{figure}[ht] 
    \centering 
\resizebox {0.8\linewidth} {!} {
\begin{tikzpicture}[y=.5cm, x=.7cm,font=\footnotesize\sffamily]
	\node[below=0.5cm] at (5,0) {number of training epochs};
	\node[rotate=90, above=0.5cm] at (-1.25, 5) {$F_1$ score and Training loss};
    \begin{axis}[ymax=1.0]
\addplot[dashed, deepred, mark=none, mark size=1pt] table[x=x,y=y] {data1.dat};

\addplot [name path=upper,draw=none] table[x=x,y expr=\thisrow{y}+1.0*\thisrow{err}] {data1.dat};
\addplot [name path=lower,draw=none] table[x=x,y expr=\thisrow{y}-1.0*\thisrow{err}] {data1.dat};
\addplot [fill=deepred!10] fill between[of=upper and lower];
\addplot[dashed, deepblue, mark=none, mark size=1pt] table[x=x,y=y] {data3.dat};

\addplot [name path=upper,draw=none] table[x=x,y expr=\thisrow{y}+1.0*\thisrow{err}] {data3.dat};
\addplot [name path=lower,draw=none] table[x=x,y expr=\thisrow{y}-1.0*\thisrow{err}] {data3.dat};
\addplot [fill=deepblue!10] fill between[of=upper and lower];
\addplot[deepred, mark=none, mark size=1pt] table[x=x,y=y] {data2.dat};

\addplot [name path=upper,draw=none] table[x=x,y expr=\thisrow{y}+1.0*\thisrow{err}] {data2.dat};
\addplot [name path=lower,draw=none] table[x=x,y expr=\thisrow{y}-1.0*\thisrow{err}] {data2.dat};
\addplot [fill=deepred!10] fill between[of=upper and lower];
\addplot[deepblue, mark=none, mark size=1pt] table[x=x,y=y] {data4.dat};
\addplot [name path=upper,draw=none] table[x=x,y expr=\thisrow{y}+1.0*\thisrow{err}] {data4.dat};
\addplot [name path=lower,draw=none] table[x=x,y expr=\thisrow{y}-1.0*\thisrow{err}] {data4.dat};
\addplot [fill=deepblue!10] fill between[of=upper and lower];
    \end{axis}
	\begin{scope}[shift={(2.5,10.8)}]
	\draw [deepred] (0,0) -- (0.5,0)
		node[right]{$F_1$ with init};
    \draw[yshift=-\baselineskip, deepblue] (0,0) --
		 (0.5,0)
		node[right]{$F_1$ w/o init};
	\draw[dashed, yshift=-1*2\baselineskip, deepred] (0,0) -- (0.5,0)
		node[right]{Loss w/ init};
	\draw[dashed, yshift=-1*3\baselineskip, deepblue] (0,0) -- (0.5,0)
		node[right]{Loss w/o init};
	\end{scope}
\end{tikzpicture}
}
    \caption{\small Comparison between random initialization and taking numbers for initialization on entity id tokens. Solid lines are $F_1$ scores on DocRED Dev set, and dashed lines are training loss on DocRED Train set. We omit the cross entropy loss higher than 1.0 for better visualization. }
    \label{fig:init_extra_id}
\end{figure}

As we mentioned before, we initialize symbols that represent entity ids by numbers, since these symbols are not trained in the pre-training phase, so the model can not recognize their meaning.
Alternatively, the model can learn from scratch during fine-tuning. However, we find the cold-start costs time and makes the training unstable. Note this strategy is very similar to position embedding used in standard Transformer-based models.
We also try to initialize the relation embedding by averaging word embeddings of their lexical forms, whereas we find it does not influence the performance significantly. 

Figure~\ref{fig:init_extra_id} shows that the initialized method converges fast and achieves a higher performance. The training loss of the first ten epochs illustrates a big gap between the cold-start and warm-start methods. That demonstrates the effectiveness of our warm start strategy.

\subsection{Backbones}
\label{sec:backbone}

\begin{table*}[!t]
\centering
\small
\begin{tabular}{lccccc}
    \toprule
    \bf Hyperparameter & \multicolumn{2}{c}{\bf DocRED} & \bf CDR & \bf GDA & \bf SciREX \\
    \midrule
    Backbone & T5 & BART/LED & BioBART & BioBART & LED \\
    Batch size & \multicolumn{2}{c}{4} & 4 & 4 & 32 \\
    Training epochs & \multicolumn{2}{c}{40} & 40 & 10 & 40 \\
    Learning rate & 1e-4 & 3e-5 & 2e-5 & 2e-5 & 5e-5 \\
    Warmup ratio & \multicolumn{2}{c}{0.06} & 0.1 & 0.15 & 0.06 \\
    Max input length & \multicolumn{2}{c}{1024} & 1024 & 1024 & 4096 \\
    Beam size & \multicolumn{2}{c}{4} & 1 & 1 & 1 \\
    \bottomrule
\end{tabular}
\caption{\small Hyperparameters used for each dataset.}
\label{tab:hyperparam}
\end{table*}

\begin{table}[!htbp]
    \centering\small
    \begin{tabular}{ccc}
    \toprule
    \bf Backbone & \bf Ign $\bf F_1$ & $\bf F_1$ \\
    \midrule
    T5$_{\textrm{large}}$ & 56.94 & 58.95 \\
    BART$_{\textrm{large}}$ & 56.96 & 59.22 \\
    LED$_{\textrm{large}}$ & 57.04 & 59.10 \\
    \bottomrule
    \end{tabular}
    \caption{\small Results of DORE using different backbones on the development set of DocRED.}
    \label{tab:backbone}
\end{table}

In principle, the method we proposed can be adapted to any pre-trained generative language models. We verify the supposition by changing the backbone with the same symbolic and ordered sequence representation and diagonal negative sampling. From Table~\ref{tab:backbone} we can see that T5, BART, or LED can achieve comparable results with our simple constrained decoding strategy, which proves the generalization ability of DORE.


\subsection{Hyperparameters}
\label{sec:appendix-hyperparam}

In Table~\ref{tab:hyperparam}, we list the hyperparameters used when training the model for each dataset. When beam size = 1, we use greedy search decoding. When beam size = 4, we use beam search decoding, and tune the length penalty $\alpha = \{0.2, \dots, 2.0\}$ with a step size of 0.2.

\subsection{Visualization details}
\label{sec:appendix-vis}
In detail, we use the attention scores from the last decoder layer of T5$_{\text{large}}$, and then we sum all attention heads. We conduct this visualization with our ``DORE + negative sampling$_\text{diagonal}$ + distant'' model, and the attention score of a triple is computed by adding up the attention scores of all its member tokens. Also, we do not consider the decoder start token ``<BOS>''. 

In this way, we compute the attention score of previously generated triples for each time that the model predict the relation type. As a result, each triple will point to a triple before it as we shown in Figure~\ref{fig:vis}. 



\subsection{Consistent Optimum}
\label{sec:appendix-optimum}
\begin{theorem}
Let $\bS^*=\tau(\bR^*)= \argmax_{\bS \in \mathbb{S}}  P(\bS|\bD, \bE)$, then we have $\bR^* = \argmax_{\bR \in \mathbb{R}} P(\bR|\bD,\bE)$.
\label{theorem:2}
\end{theorem}

\paragraph{Proof}
Since $\bS^*= \argmax_{\bS \in \mathbb{S}} P(\bS|\bD, \bE)$, so for any $\bS \in \mathbb{S}$, we have $P(\bS^*|\bD,\bE) - P(\bS | \bD, \bE) \ge 0$. According to the eq.~\eqref{eq:seq2seq}, we can rewrite the formulation. 
\begin{align}
    &P(\bR^* | \bD,\bE) - P(\bR | \bD, \bE), \\
    =&P(\bS^* | \bD, \bE) - P(\tau(\bR) | \bD, \bE), \\
    \ge& 0.
\end{align}

\end{document}